\documentclass[11pt]{article}

\usepackage[preprint]{acl}      

\usepackage{times}
\usepackage{latexsym}
\usepackage[T1]{fontenc}
\usepackage[utf8]{inputenc}
\usepackage{microtype}

\usepackage{amsmath, amssymb, amsthm}
\usepackage{booktabs}
\usepackage{graphicx}
\usepackage{multirow}
\usepackage{array}
\usepackage{colortbl}
\usepackage{xcolor}

\usepackage{url}
\usepackage{hyperref}
\hypersetup{
  colorlinks=true,
  citecolor=blue,
  linkcolor=blue,
  urlcolor=cyan,
  pdftitle={A Unified Per-Token Gating Family for On-Policy Distillation: FKL/RKL Mixing with Multi-Channel and Bias Coefficients},
  pdfauthor={Suwan Wu, Yumeng Lin, Pengcheng Yuan, Xiaolong Jiang},
  pdfsubject={Knowledge distillation; on-policy distillation; per-token KL gating},
  pdfkeywords={knowledge distillation, on-policy distillation, forward KL, reverse KL, per-token gating, TweetEval},
}

\newcommand{\todi}{ToDi}
\newcommand{\eopd}{EOPD}

\newcommand{\fkl}{\ensuremath{\mathcal{L}_{\mathrm{FKL}}}}
\newcommand{\rkl}{\ensuremath{\mathcal{L}_{\mathrm{RKL}}}}
\newcommand{\hpt}{\ensuremath{h_t}}
\newcommand{\ux}{\ensuremath{u(x)}}
\newcommand{\gapt}{\ensuremath{\mathrm{gap}_t}}
\newcommand{\lamt}{\ensuremath{\lambda_t}}
\newcommand{\todial}{gap-only}
\newcommand{\eopdal}{entropy-only}

\title{A Unified Per-Token Gating Family for On-Policy Distillation:\\
       FKL/RKL Mixing with Multi-Channel and Bias Coefficients}

\author{%
  Suwan Wu\textsuperscript{1,*} \quad Yumeng Lin\textsuperscript{1,2,*} \quad
  Pengcheng Yuan\textsuperscript{1} \quad Xiaolong Jiang\textsuperscript{1} \\
  \textsuperscript{1}Xiaohongshu Inc. \quad \textsuperscript{2}Tianjin University \\
  \texttt{\{wusuwan, linyumeng, yuanpengcheng, laige\}@xiaohongshu.com} \\
  \texttt{lym619@tju.edu.cn}
}

\begin{document}
\maketitle

\begingroup
\renewcommand{\thefootnote}{}
\footnotetext{* Suwan Wu and Yumeng Lin contributed equally. Accepted at the Findings of the 2026 Conference on Empirical Methods in Natural Language Processing (EMNLP 2026 Findings).}
\endgroup
\addtocounter{footnote}{0}

\begin{abstract}
Per-token gating of forward/reverse KL losses has become a standard technique for on-policy knowledge distillation (OPD), but existing methods such as \eopd{} \citep{jin2026entropy} and \todi{} \citep{jung2025todi} each fix a single gating signal and a single gating direction, and the two have never been compared directly. We introduce a four-coefficient parameterization $\lamt = \sigma(a \cdot \hpt + b \cdot \ux + c + d \cdot \gapt)$ in which direction-aligned proxies of \eopd{} and \todi{} appear as one-dimensional (1D) restrictions, and which adds multi-channel composition and an explicit bias as further degrees of freedom. On TweetEval \citep{barbieri2020tweeteval} emotion and hate, with a Qwen3-32B teacher and a Qwen3-4B student, configurations in the full family reach higher accuracy than the matched-magnitude single-channel (entropy-only / gap-only) 1D restrictions in 33 of 36 comparable cells, and a 26-cell mean-match isolation experiment places dynamic gating ahead of effective-KL-matched static baselines in 19 of 26 cells. Because cells share training data, models, and parameter substructure, we report both counts as exploratory aggregate directional evidence rather than as independent hypothesis tests. Targeted three-seed paired replications of the nine headline comparisons singled out by that sweep --- including a third task, offensive --- are directionally consistent, but individually smaller than the single-seed estimates and not significant at $n{=}3$. We therefore present the parameterization primarily as a shared coordinate system for comparing per-token gating designs in short-output classification OPD.
\end{abstract}

\section{Introduction}
\label{sec:intro}

Knowledge distillation (KD) from large language models (LLMs) to smaller student models typically combines forward Kullback--Leibler (FKL) and reverse KL (RKL) losses to balance two complementary objectives: FKL drives the student to cover the teacher's full output distribution (mode-covering), while RKL drives mode-seeking on the teacher's high-probability regions \citep{kim2016sequencelevel, gu2024minillm}. The relative weighting of these two losses --- captured by a single scalar $\lambda \in [0, 1]$ in standard formulations --- determines the student's distillation regime.

Recent work has argued that \textbf{fixed $\lambda$ is suboptimal} and that per-token dynamic gating $\lamt \in [0, 1]$ helps \citep{jin2026entropy, jung2025todi}. The two dominant approaches differ in their choice of gating signal:

\begin{itemize}
\item \textbf{\eopd{}} \citep{jin2026entropy} uses \emph{teacher entropy} \hpt{}: high entropy tokens (where the teacher is uncertain) receive more FKL weight to encourage student exploration.
\item \textbf{\todi{}} \citep{jung2025todi} uses \emph{teacher--student disagreement} (the log-ratio of teacher to student probability, computed for every vocabulary entry): entries on which the teacher places more mass than the student receive more FKL weight, raising the student's probability there.
\end{itemize}

Both lines of work report improvements over static baselines under their respective protocols, but two questions remain: why per-token gating helps, and which signal is preferable for a given task. Since each fixes a single signal and a single gating direction, it has not been tested whether their improvements come from the signal choice, from gating in general, or from hyperparameter tuning --- and the two cannot be compared directly, as they operate in different parameter subspaces with different conventions.

\paragraph{A unified analysis framework.}
Rather than proposing a new gating method, we introduce a parametric family that turns the choice between \eopd{}, \todi{}, and their variants from a discrete method selection into a continuous point in a shared parameter space:
\begin{equation}
\lamt = \sigma\bigl( a \cdot \hpt + b \cdot \ux + c + d \cdot \gapt \bigr),
\label{eq:family}
\end{equation}
where \hpt{} is the per-token normalized teacher entropy, \ux{} is the per-sample prompt-level teacher entropy, $\gapt = 1 - p_{\text{student}}(y_t^{\text{teacher-top1}})$ is the per-token teacher--student disagreement, and $c$ is a bias term. The per-token mixture is $\mathcal{L}(t) = \lamt \rkl(t) + (1 - \lamt) \fkl(t)$.

Under this parameterization, direction-aligned proxies of the prior methods appear as 1D restrictions:
\begin{itemize}
\item \textbf{\eopd{}} corresponds to $(-\beta, 0, 0, 0)$ --- only token-level entropy, no bias, with the sign reversed by the original ``high entropy $\to$ more FKL'' semantics.
\item \textbf{\todi{}} corresponds to $(0, 0, 0, -\beta)$ --- only teacher--student disagreement, no bias. The sign is negative because \todi{}'s weight multiplies FKL whereas our \lamt{} multiplies RKL (Section~\ref{sec:special-cases}).
\end{itemize}
As we detail in Section~\ref{sec:special-cases}, these 1D restrictions are \emph{structurally aligned proxies} for the original methods within our convex-mixture family, not faithful reproductions of the published algorithms; all comparisons in this paper should be read accordingly.

\paragraph{What this framework enables.}
Our central contribution is this unified parameterization. It reframes two prior questions as questions about points in a shared parameter space:
\begin{enumerate}
\item \textbf{Which signal helps which task?} The framework exposes four signal coefficients independently, so the grid analysis in Section~\ref{sec:main} can ask whether different tasks prefer different channels rather than assuming one gating signal a priori. Single-restriction methods cannot represent this trade-off.
\item \textbf{Do dynamic gates contribute beyond what a constant $\lambda$ captures?} The framework allows a clean isolation protocol --- training a static baseline at the emergent effective KL ratio of each dynamic configuration --- to disentangle dynamic structure from average KL weighting. Section~\ref{sec:mean-match} reports this 26-cell isolation experiment.
\end{enumerate}

\paragraph{Empirical contributions.}
We instantiate the framework with student Qwen3-4B and teacher Qwen3-32B \citep{qwen3} and evaluate on TweetEval emotion (4-way classification, $n_{\mathrm{test}}=1421$) and hate (binary, $n_{\mathrm{test}}=2970$) \citep{barbieri2020tweeteval}, plus a third task, offensive (binary, $n_{\mathrm{test}}=860$), added for seed-robustness replication (Section~\ref{sec:seeds}). Our sweeps span 13 OPD configurations per task across the $(a, b, c, d)$ space (Section~\ref{sec:main}: $G_1$--$G_5$, $H_1$--$H_4$, $HG_{d\pm 2, \pm 4}$), each compared against the aligned single-coefficient restrictions corresponding to \todi{} and \eopd{} at the same signal magnitude.

Three findings emerge:

\textbf{Finding 1 --- Structural coverage.} Hate's best configuration in the sweep, $G_4 = (0, 4, -1, 0)$, uses the \ux{} channel, \textbf{structurally absent in both \todi{} and \eopd{}}. Across the 26 (config $\times$ task) cells the two restrictions give $52$ potential comparisons, of which 10 are ToDi-N/A and 6 are EOPD-N/A --- configurations a matched-magnitude restriction cannot represent by construction --- leaving the $36$ comparable cells used below. The parameterization strictly extends the union of the two 1D restrictions.

\textbf{Finding 2 --- Aggregate directional advantage at matched magnitude.} Across the 36 comparable cells (emotion 20 + hate 16), the full family beats the matched 1D restriction in \textbf{33 cells (91.7\%)}, with one unfavorable cell outside the sampling-SE reference band that Section~\ref{sec:granularity} identifies as a granularity mismatch. Because cells share data, models, and parameter substructure, we report this count as exploratory aggregate directional evidence and attach no significance to individual gaps.

\textbf{Finding 3 --- The aggregate gain is not explained by the effective KL ratio.} In 26 isolation experiments against mean-matched static baselines sharing the same effective KL ratio (Section~\ref{sec:mean-match}), dynamic gating is ahead in \textbf{19 of 26 configurations}, none below the negative $1\sigma$ sampling-SE reference band --- so the per-token structure appears to carry information a constant $\lambda$ at the same ratio does not.

Section~\ref{sec:seeds} then quantifies how much of the per-cell magnitude is seed noise: nine headline comparisons singled out by the single-seed sweep, spanning both axes and all three tasks, were re-run with three seeds on \emph{both} sides. All three-seed means are smaller than the single-seed estimates, eight of nine remain directionally positive, and none is individually significant at $n{=}3$; we consequently state all empirical claims at the group level.

\section{Related Work}
\label{sec:related}

We organize prior work along three threads: (i) KD losses with mixed forward/reverse KL, (ii) per-token gating methods (the most direct comparators), and (iii) on-policy distillation infrastructure. The two comparators come from different training regimes: \eopd{} is defined inside the on-policy distillation framework of \citet{agarwal2024onpolicy} (summarized in Appendix~\ref{app:repro}), whereas \todi{} was proposed for offline distillation on a fixed instruction-tuning corpus.

\subsection{Mixed FKL/RKL Distillation}

Sequence-level distillation \citep{kim2016sequencelevel} extended KD to autoregressive sequence models by training the student on teacher-generated sequences, with most subsequent work using forward KL $D_{\mathrm{KL}}(p_{\text{teacher}} \| p_{\text{student}})$ --- a mode-covering objective. MiniLLM \citep{gu2024minillm} noted that mode-covering produces high-quality but generic outputs and proposed reverse KL $D_{\mathrm{KL}}(p_{\text{student}} \| p_{\text{teacher}})$ as a mode-seeking alternative for instruction-following. A common practical compromise is a fixed mixture $\mathcal{L} = \lambda \rkl + (1 - \lambda) \fkl$, with $\lambda$ a hyperparameter --- typically $0.5$. Adaptive variants have been explored at the sequence or batch level using moving statistics of teacher--student divergence; such schemes operate at sample granularity and do not exploit per-token signals.

\subsection{Per-Token Gating: EOPD and ToDi}

\textbf{\eopd{}} \citep{jin2026entropy} introduced an entropy-driven per-token gating mechanism: a hard switch on teacher entropy,
\begin{equation}
\mathcal{L}_t^{\eopd{}} = \mathcal{L}_t^{\text{OPD}} + \alpha \cdot \mathbb{I}[H_t^{\mathrm{te}} > \tau] \cdot \fkl(t),
\label{eq:eopd}
\end{equation}
where $H_t^{\mathrm{te}}$ is the teacher's unnormalized token-level entropy over the vocabulary, and tokens on which the teacher is uncertain receive an additional FKL term that preserves the teacher's distributional diversity; the original work uses $\tau = 0.8$ and $\alpha = 1$. The structure is \textbf{additive} rather than convex, and the gating signal is \textbf{teacher entropy} alone. Our \hpt{} (Section~\ref{sec:coef}) is a bounded top-$K$ analogue of $H_t^{\mathrm{te}}$, normalized to $[0,1]$.

\textbf{\todi{}} \citep{jung2025todi} introduced a convex mixture whose weight is computed \emph{per vocabulary entry} $v_i$ from the teacher--student log-ratio:
\begin{equation}
\alpha_{t,i} = \mathrm{sg}\!\left[\sigma\!\left(\beta \cdot \log \tfrac{p(v_i \mid \mathbf{y}_{<t}, \mathbf{x})}{q_\theta(v_i \mid \mathbf{y}_{<t}, \mathbf{x})}\right)\right],
\label{eq:todi}
\end{equation}
with $D^{(t,i)}_{\todi{}} = \alpha_{t,i} D^{(t,i)}_{\mathrm{FKL}} + (1 - \alpha_{t,i}) D^{(t,i)}_{\mathrm{RKL}}$ summed over positions and vocabulary entries. \todi{} targets a different mechanism: entries where the teacher places more mass than the student ($p > q_\theta$) receive more \emph{FKL}, raising the student's probability there, while over-estimated entries receive more RKL. Its signal, \textbf{teacher--student disagreement}, is closely related to our $\gapt$ --- both grow as the student under-estimates a token the teacher favours --- but enters with the opposite sign, since $\alpha_{t,i}$ multiplies FKL whereas \lamt{} multiplies RKL.

\textbf{Direct empirical comparison between \eopd{} and \todi{} has not been reported}: they operate in different parameter subspaces (entropy vs.\ disagreement), use different scaling conventions, and were evaluated on different datasets, models and metrics. Our framework (Section~\ref{sec:coef}) places direction-aligned proxies of both inside one four-coefficient family, enabling a controlled comparison at matched magnitude: \eopd{} maps to $(-\beta, 0, 0, 0)$ and \todi{} to $(0, 0, 0, -\beta)$, both negative because each prior method routes its signal towards FKL whereas \lamt{} weights RKL. Section~\ref{sec:vs-prior} reports the 26-cell comparison and documents where the proxies depart from the published algorithms.

\section{Method: Parametric Family and Aligned Restrictions of Prior Methods}
\label{sec:method}
\label{sec:coef}

\subsection{Parametric Family}

Our family is the four-coefficient parameterization of per-token \rkl{}/\fkl{} mixing (Equation~\ref{eq:family}). The three input signals are: \textbf{token-level teacher entropy} $\hpt \in [0, 1]$ (normalized entropy of the teacher's top-$K$ predictions at token $t$; high \hpt{} indicates the teacher is uncertain); \textbf{sample-level prompt entropy} $\ux \in [0, 1]$ (teacher entropy on the first decoded token after the prompt, computed once per sample); and \textbf{teacher--student disagreement} $\gapt = 1 - p_{\text{student}}(y_t^{\text{teacher-top1}}) \in [0, 1]$ (probability mass the student assigns to a token other than the teacher's top-1 prediction). All three signals are non-negative and signed consistently: a larger value means ``the student needs to learn more teacher mode at this token / sample.''

\subsection{Coefficient Patterns Observed in the Sweep}

Two recurring patterns and one task-dependent observation emerge from the Section~\ref{sec:main} sweep. All three are descriptive summaries of a single-seed sweep on two tasks, and Section~\ref{sec:seeds} shows that individual cell magnitudes are seed-sensitive, so none should be read as an established rule.

\textbf{Pattern 1: $c < 0$ (negative bias).} The best configuration on each task uses $c = -1$, while the unbiased $G_1 = (2, 2, 0, 0)$ is behind on both. Negative bias shifts the prior on \lamt{} below $0.5$, favoring \fkl{} on indifferent tokens; the effect is clear on hate and within noise on emotion (Section~\ref{sec:taskmirror}).

\textbf{Pattern 2: $(a, b)$ asymmetry.} The best configuration on each task zeroes one of $a, b$ and amplifies the other ($G_4$ on hate, $G_5$ on emotion), with the symmetric all-on baselines $G_2, G_3$ behind both --- consistent with sigmoid saturation, since $|a|+|b|$ in the same direction saturates the gate and loses discriminative power. As with Pattern 1, the emotion side of this contrast lies within the sampling-SE reference band.

\textbf{Observation: the $d$ direction is task-dependent.} The gap signal can be same-sign as the $abc$ signal (reverse-gap, more RKL on disagreement tokens) or opposite (forward-gap, more FKL). Both directions improved a weak $abc$ baseline in this sweep, whereas stacking $d$ on the task-best $abc$ helped only when granularities matched (Sections~\ref{sec:d-on-g2}--\ref{sec:granularity}).

\subsection{Aligned 1D Restrictions of Prior Methods}
\label{sec:special-cases}

Our family contains two single-coefficient restrictions isolating the signals used by \todi{} and \eopd{}. We name them by the channel retained --- \textbf{\todial{}} $(0,0,0,d)$ and \textbf{\eopdal{}} $(-a,0,0,0)$ --- rather than by method name, since only one sign branch of the former matches the published method.

\paragraph{\todial{} restriction and \todi{}'s direction.}
\todi{}'s log-ratio signal and our prob-diff $\gapt$ grow together (both increase as the student under-estimates a token the teacher favours), but \todi{}'s weight multiplies FKL while \lamt{} multiplies RKL. Translating Equation~\ref{eq:todi} into our convention gives $\lamt = 1 - \alpha_t = \sigma(-\beta \log (p/q))$, so within our family \todi{}'s gating direction is
\begin{equation}
\boxed{(a, b, c, d) = (0, 0, 0, -\beta)}, \quad \beta > 0.
\label{eq:todi-mapping}
\end{equation}
Our \todial{} restriction $(0,0,0,d)$ therefore reproduces \todi{}'s direction when $d < 0$ and reverses it when $d > 0$; the sweep contains both branches, as each restriction matches the sign of its paired configuration. Even for $d<0$ it is a structural analogue rather than a reproduction; Section~\ref{sec:proxy-caveat} enumerates the differences.

\paragraph{\eopdal{} restriction and \eopd{}'s direction.}
\eopd{} is additive (OPD plus a conditional FKL add-on, Equation~\ref{eq:eopd}), while our family uses a convex mixture, so we do not reproduce the additive form. Instead we use a \textbf{convex relaxation} that preserves the qualitative behavior ``high $\hpt$ $\to$ more FKL''. Every \eopdal{} run in our sweep sets the bias to zero, so the restriction is
\begin{equation}
\lamt^{\eopdal{}} = \sigma(-\beta \hpt),
\quad\text{i.e.}\quad
\boxed{(-\beta, 0, 0, 0)},
\label{eq:eopd-mapping}
\end{equation}
with $\beta \in \{2,4,8\}$: high $\hpt$ drives $\lamt$ towards $0$ (FKL-heavy), matching \eopd{}'s direction. Two caveats follow. With zero bias the gate is \emph{monotone and soft}, anchored at $\lamt = 0.5$ for $\hpt = 0$ and bounded above by it; \eopd{}'s hard threshold would need a positive bias $c=\tau$, since $\sigma(\tau - \beta\hpt) \to \mathbb{I}[\hpt < \tau/\beta]$ as $\beta \to \infty$ only for $\tau>0$, and we did not sweep $\tau$. The negative sign on $a$ is fixed by \eopd{}'s ``high entropy $\to$ FKL'' semantics, so --- unlike the gap channel --- every \eopdal{} restriction in our sweep lies on \eopd{}'s side of the sign convention.

\paragraph{What the comparison does and does not test.}
The empirical comparison in Section~\ref{sec:vs-prior} tests a structural claim inside a single implementation: multi-coefficient combinations outperform single-coefficient 1D restrictions of the same family at matched signal magnitude. The parameterization makes the design space explicit: the \todial{} restriction is $(0, 0, 0, d)$ (\todi{}'s direction being $d<0$), the \eopdal{} restriction is $(-a, 0, 0, 0)$, and both fix $c = 0$; our family exposes multi-channel composition and explicit bias as additional degrees of freedom. It is \emph{not} a reproduction of the published \eopd{} or \todi{} systems, and results here should not be read as a ranking of those systems (Section~\ref{sec:proxy-caveat}).

\section{Main Results: Grid Analysis}
\label{sec:main}

We instantiate the $(a, b, c, d)$ family with student Qwen3-4B and teacher Qwen3-32B and evaluate on TweetEval held-out evaluation splits: emotion (4-way classification, $n=1421$) and hate (binary, $n=2970$). Training: 100 steps (emotion) / 200 steps (hate), batch size 72, learning rate $10^{-6}$, ZMQ-based on-policy distillation. All accuracy values reported in this section come from a single seed per configuration; Section~\ref{sec:seeds} replicates the headline comparisons with three seeds.

\paragraph{Configuration sweep protocol.}
The $(a, b, c, d)$ grid (13 configurations per task) was chosen prior to evaluating any specific configuration on the held-out split; the same configurations are used in this section's family analysis and in Section~\ref{sec:vs-prior}'s comparison against aligned restrictions. Reported aggregate statistics (win-rate counts) treat each configuration as a unit but do \emph{not} assume that configurations are statistically independent (Section~\ref{sec:vs-prior}).

\subsection{Grid Design}

We explore the $(a, b, c, d)$ space along three axes:
\begin{itemize}
\item \textbf{$abc$ grid} (5 configs with $d = 0$): $G_1 = (2, 2, 0, 0)$, $G_2 = (4, 4, -1.5, 0)$, $G_3 = (8, 8, -3, 0)$, $G_4 = (0, 4, -1, 0)$, $G_5 = (4, 0, -1, 0)$. Varies the balance between the sample (\ux{}) and token (\hpt{}) channels at three magnitudes plus two single-channel configurations.
\item \textbf{$d$ on a weak baseline ($H$ family)} (4 configs with $d \in \{\pm 2, \pm 4\}$ stacked on $G_2$): $H_1$--$H_4$. Tests whether the \gapt{} channel improves a non-saturated $abc$ baseline.
\item \textbf{$d$ on the task-best $abc$ ($HG$ family)} (4 configs each): emotion $HG_{5,d\pm 2,\pm 4} = (4, 0, -1, \pm d)$ on the $G_5$ base; hate $HG_{4,d\pm 2,\pm 4} = (0, 4, -1, \pm d)$ on the $G_4$ base.
\end{itemize}
In total: 13 OPD configurations per task = 26 (config $\times$ task) cells.

\subsection{$abc$ Grid: Task-Dependent on Hate, Flat on Emotion}
\label{sec:taskmirror}

\begin{figure}[t]
\centering
\includegraphics[width=\columnwidth]{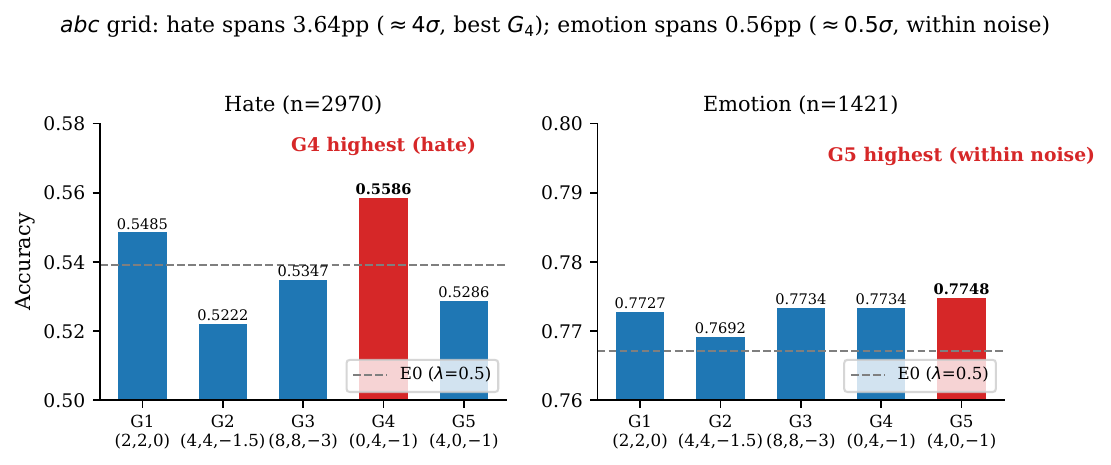}
\caption{$abc$ grid accuracy. The hate-best $G_4 = (0, 4, -1)$ and emotion-best $G_5 = (4, 0, -1)$ are coordinate swaps with identical $c=-1$. The hate grid spans 3.64pp ($\approx 4\sigma$); the emotion grid spans 0.56pp ($\approx 0.5\sigma$), i.e.\ the emotion ordering is inside the sampling-SE reference band and we draw no channel preference from it.}
\label{fig:task-mirror}
\end{figure}

Figure~\ref{fig:task-mirror} shows the $abc$ grid accuracy on both tasks. On \textbf{hate}, the sweep selects $G_4 = (0, 4, -1)$: sample-level \ux{} only, with negative bias (ACC $=0.5586$). The accuracy range across the grid is \textbf{3.64pp}, i.e.\ $4\sigma$ against a \emph{heuristic test-set sampling-SE reference band} ($\sigma$ = binomial SE of a single model's accuracy at $n=2970$; it is not the SE of a paired difference, and it ignores training-seed variability, so we use it only as an order-of-magnitude reference). On \textbf{emotion}, the sweep selects $G_5 = (4, 0, -1)$: token-level \hpt{} only, with negative bias (ACC $=0.7748$) --- but the entire grid spans only \textbf{0.56pp} ($\approx 0.5\sigma$ at $n=1421$).

The two selected configurations are coordinate swaps ($G_4$: $a=0, b=4$; $G_5$: $a=4, b=0$) with identical $c=-1$. We nevertheless \textbf{claim no task-conditional channel preference on emotion}: with the whole emotion grid inside $0.5\sigma$, the emotion-best cell is indistinguishable from noise and the symmetry is a descriptive coincidence of this sweep. Only the narrower statement is supported: \emph{on hate}, the configuration putting all weight on \ux{} was strongest, over a range exceeding the sampling band. Establishing a task-conditional preference would need multi-seed replication of the full grid, which we did not run (Limitations).

\subsection{$d$ on $G_2$ (Weak Baseline)}
\label{sec:d-on-g2}

\begin{table}[t]
\centering
\small
\setlength{\tabcolsep}{4pt}
\begin{tabular}{lcccc}
\toprule
$\Delta$ vs $G_2$ (pp) & $d{=}{+}2$ & $d{=}{+}4$ & $d{=}{-}2$ & $d{=}{-}4$ \\
\midrule
Hate    & \textbf{+3.27} & +1.89 & +1.35 & +1.45 \\
Emotion & +0.63 & \textbf{+1.55} & +0.42 & +1.41 \\
\bottomrule
\end{tabular}
\caption{Adding $d$ to the weak $G_2$ baseline: positive in 8/8 settings of this single-seed sweep; max $+3.27$pp on hate.}
\label{tab:d-on-g2}
\end{table}

Adding $d$ to the weak $G_2$ baseline improved accuracy in \textbf{all 8 settings of this sweep} (Table~\ref{tab:d-on-g2}). We read this as the \gapt{} channel being \emph{beneficial across the settings we tested} on a non-saturated baseline; with two tasks, one seed, and one base configuration, it does not establish that the channel is universally informative.

\subsection{$d$ on the Task-Best $abc$: Granularity-Conditional}
\label{sec:granularity}

On emotion, adding $d$ to the token-best $G_5$ produces $HG_{5,d+4} = 0.7868$ (the highest emotion cell in the sweep), consistent with matched granularity (token-level $abc$ plus token-level gap). On hate, adding $d$ to the sample-best $G_4$ degrades performance in all four settings: $G_4$'s gating decision is per-sample (all tokens of a response share one $\lambda$), so injecting a per-token \gapt{} creates a granularity mismatch (Figure~\ref{fig:granularity} in Appendix~\ref{app:extra}). This mismatch interpretation is the one place where the sweep produces a cell outside the sampling-SE reference band in the unfavorable direction (Section~\ref{sec:hate-table}); under three-seed replication that cell shrinks to within noise (Section~\ref{sec:seeds}), so the mechanism should be regarded as a hypothesis rather than a demonstrated effect.

\subsection{Dynamic vs.\ Mean-Matched Static (Isolation)}
\label{sec:mean-match}

For each of the 13 OPD configurations $\times$ 2 tasks $=26$ dynamic runs, we compute the emergent training-time \lamt{} mean and train a static baseline with $\mathrm{rkl\_ratio} = \mathbb{E}[\lamt]$ on the same task. This controls for the trivial explanation that dynamic gains arise from a different effective KL ratio rather than from per-token signal exploitation.

\begin{figure*}[t]
\centering
\includegraphics[width=0.92\textwidth]{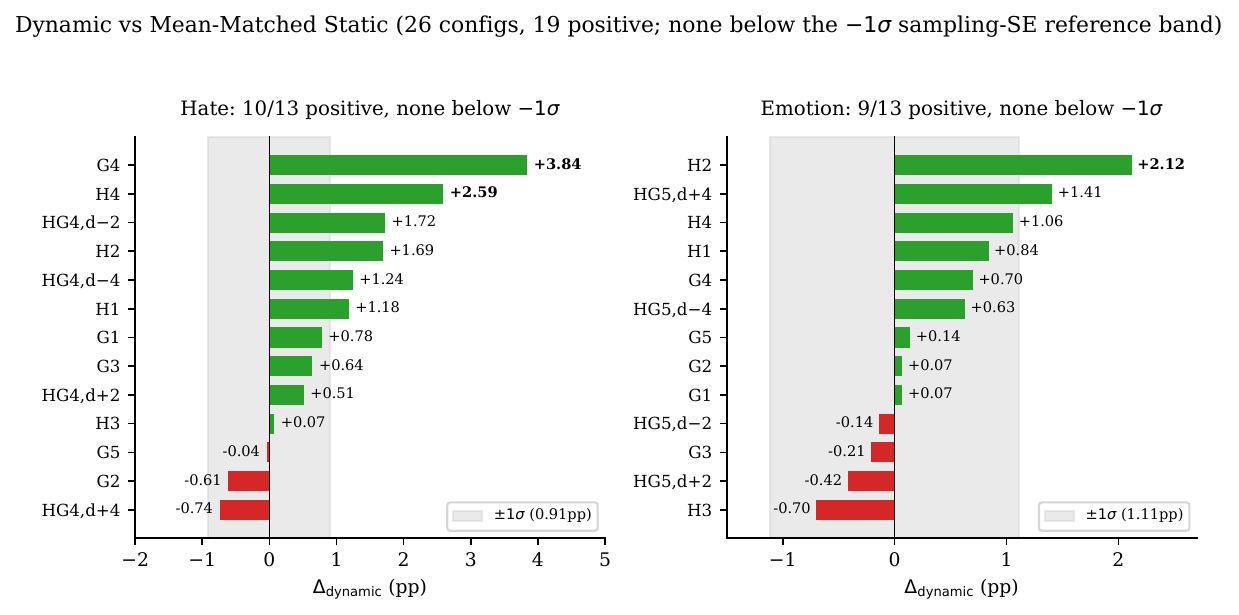}
\caption{Dynamic OPD vs.\ mean-matched static across 26 (config $\times$ task) cells. No configuration falls below the negative $1\sigma$ band and 19 of 26 cells lean positive; the cells are correlated (see the footnote in Section~\ref{sec:vs-prior}). Three cells stand out in this single-seed sweep: hate $G_4 = +3.84$pp, hate $H_4 = +2.59$pp, emotion $H_2 = +2.12$pp; the hate $G_4$ cell is $+1.43 \pm 3.09$pp under three-seed replication (Table~\ref{tab:3seed}). Gray bands denote $\pm 1\sigma$ \emph{test-set sampling} SE --- not across-seed variability, which is separately quantified in Section~\ref{sec:seeds}.}
\label{fig:mean-match}
\end{figure*}

\textbf{Directional summary}: 19/26 cells positive, with no cell below the $-1\sigma$ sampling-SE reference band (Figure~\ref{fig:mean-match}). Cells share data and substructure across configurations, so we treat this as an aggregate directional summary rather than a set of independent tests. Section~\ref{sec:seeds} replicates the highest-accuracy configuration per task on this axis with three seeds: hate $G_4$ (which is also the largest $\Delta$) and emotion $HG_{5,d+4}$ (whose largest-$\Delta$ cell is instead $H_2$).

\section{Comparison with Aligned 1D Restrictions of Prior Methods}
\label{sec:vs-prior}

\subsection{Setup}
For each of the 13 OPD configurations from Section~\ref{sec:main}, we compute the corresponding \textbf{\todial{}} and \textbf{\eopdal{}} restriction within our framework (Section~\ref{sec:special-cases}):
\begin{itemize}
\item \textbf{\todial{}}: replace $(a, b, c, d)$ with $(0, 0, 0, d)$ --- retain only the \gapt{} coefficient at the same magnitude and sign.
\item \textbf{\eopdal{}}: replace $(a, b, c, d)$ with $(-a, 0, 0, 0)$ --- retain only the \hpt{} coefficient, sign reversed by \eopd{}'s ``high entropy $\to$ FKL'' semantics.
\end{itemize}
Configurations where the aligned restriction is degenerate are marked \textbf{N/A}: ToDi-N/A when $d=0$; EOPD-N/A when $a=0$. \textbf{N/A entries are not missing data} --- they are configurations that the matched-magnitude restriction cannot represent by construction.

All runs share an identical training setup. The 13 rows for our family reuse the Section~\ref{sec:main} accuracies; the \todial{} and \eopdal{} cells are separately trained and evaluated.

\begin{figure*}[t]
\centering
\includegraphics[width=0.95\textwidth]{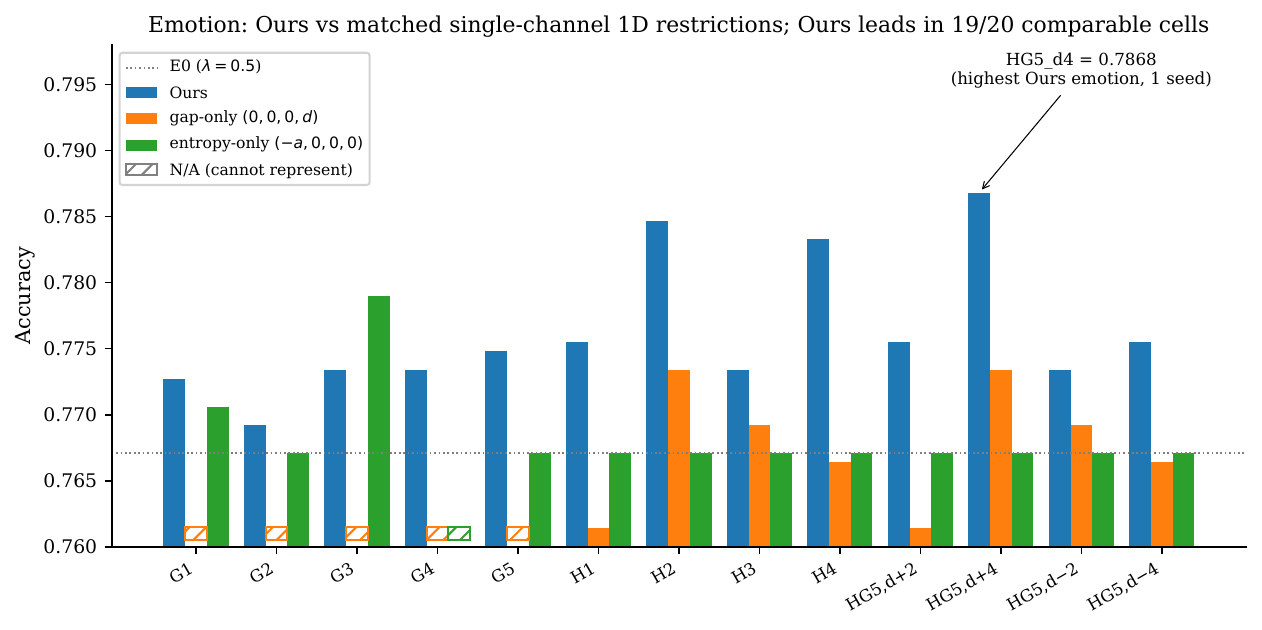}
\caption{Emotion: our family vs.\ the matched single-channel 1D restrictions, leading in 19 of 20 comparable cells. N/A bars omitted. Highest emotion cell: $HG_{5,d+4} = 0.7868$ (single seed; $0.7762 \pm 0.92$pp over three seeds, Section~\ref{sec:seeds}).}
\label{fig:vs-prior-emotion}
\end{figure*}

\begin{figure*}[t]
\centering
\includegraphics[width=0.95\textwidth]{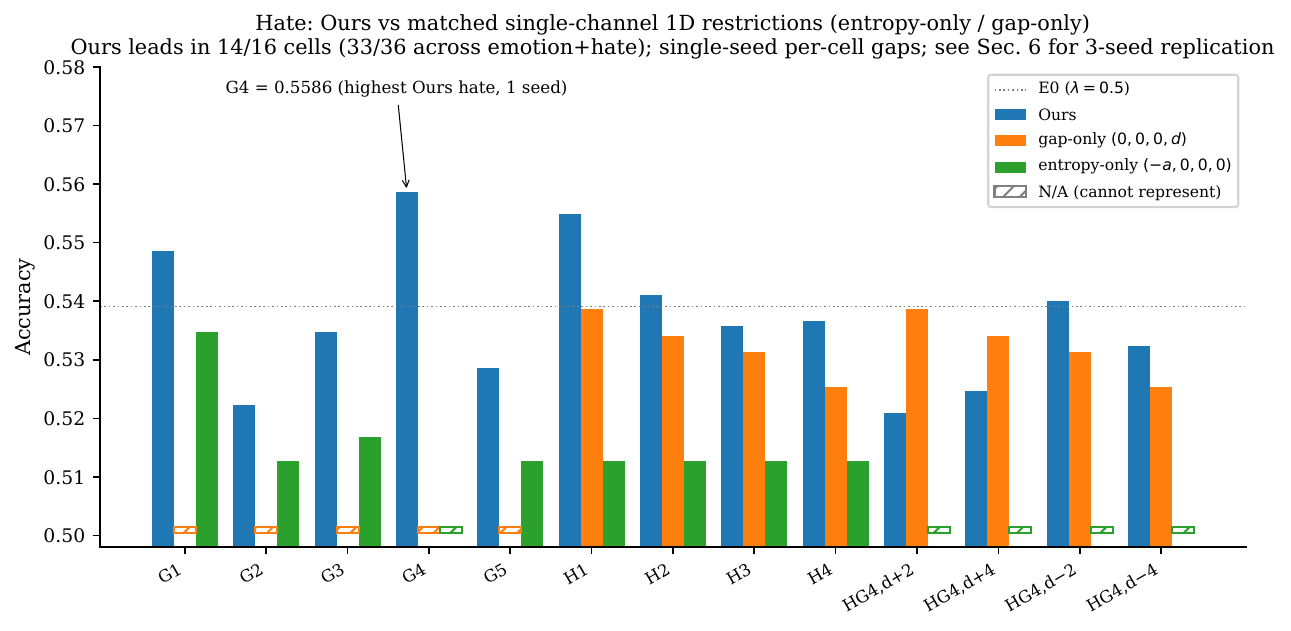}
\caption{Hate: our family vs.\ the matched single-channel 1D restrictions, leading in 14 of 16 comparable cells. The largest single-seed gap ($H_1$ vs.\ \eopdal{}$(-4)$, $+4.21$pp) shrinks to $+1.73 \pm 2.18$pp over three seeds (Section~\ref{sec:seeds}).}
\label{fig:vs-prior-hate}
\end{figure*}

\subsection{What the aligned restrictions do not capture}
\label{sec:proxy-caveat}

Because this section is the paper's main head-to-head evidence, we state its scope precisely. The comparison is between points of \emph{one} convex-mixture implementation, and the restrictions differ from the published methods in five documented ways:
(i) the \todial{} restriction uses the prob-diff gap $\gapt$ on the teacher's top-1 token instead of \todi{}'s log-ratio gate over the whole vocabulary, so it collapses a per-entry weight to one scalar per position;
(ii) it omits \todi{}'s stop-gradient on the gate, so gradients flow through the gating signal;
(iii) only its $d<0$ branch matches \todi{}'s gating direction (Section~\ref{sec:special-cases}); the $d>0$ cells are the sign-reversed variant, retained because each restriction is matched to the sign of the configuration it is paired with;
(iv) the \eopdal{} restriction replaces \eopd{}'s additive hard switch with a convex sigmoid relaxation, which changes how the FKL term enters the loss, not only when, and drops \eopd{}'s coefficient $\alpha$;
(v) neither restriction uses the $\beta$, $\tau$ or $\alpha$ tuned in the original papers --- the magnitude is instead matched to the configuration under comparison.
Consequently, results below support statements of the form ``the multi-coefficient family outperforms the entropy-only and gap-only 1D restrictions within this family under matched magnitude,'' and \emph{not} ``our method outperforms \eopd{} or \todi{}.'' We make no claim about the published systems' peak performance.

\subsection{Emotion}
\label{sec:emotion-table}

Across the 8 cells comparable to the \todial{} restriction, our family leads in 8/8 with mean $\Delta = +1.09$pp; across the 12 cells comparable to the \eopdal{} restriction, it leads in 11/12 with mean $\Delta = +0.74$pp. \textbf{Total: 19/20 cells (95\%)}. The single cell in the other direction is $G_3 = (8,8,-3,0)$ vs.\ $\eopdal{}(-8, 0, 0, 0)$ at $\Delta = -0.56$pp ($0.36\sigma$, within the reference band).

\subsection{Hate}
\label{sec:hate-table}

Across the 8 cells comparable to the \todial{} restriction, our family leads in 6/8 with mean $\Delta = +0.35$pp; the one cell outside the reference band in the unfavorable direction is $HG_{4,d+2}$ vs.\ $\todial{}(+2)$ at $-1.78$pp, which Section~\ref{sec:granularity} attributes to a granularity mismatch ($HG_4$ stacks a token-level $d$ on a sample-level $b$). Across the 8 cells comparable to the \eopdal{} restriction, our family leads in \textbf{8/8} with mean $\Delta = +2.18$pp.

\subsection{Findings}
\label{sec:best-vs-best}

Before the cell-wise findings, one aggregate check: when each side picks its strongest configuration in the explored range, our family still leads on both tasks (Table~\ref{tab:best-observed}, Appendix~\ref{app:extra}); we do not quantify that lead, since a genuine best-vs-best comparison would sweep magnitudes more widely per restriction and replicate across seeds (Limitations).

\textbf{Finding 1 --- Structural coverage.} The 26 (config $\times$ task) cells yield $52$ potential comparisons. \textbf{ToDi-N/A occurs in 10} ($G_1$--$G_5$ on both tasks, $d=0$) and \textbf{EOPD-N/A in 6} ($G_4$ on both tasks plus $HG_{4,*}$ on hate, all with $a=0$), leaving $52-10-6=36$ comparable cells (emotion 20 + hate 16). Hate's strongest configuration $G_4 = (0, 4, -1, 0)$ uses the \ux{} channel and is \textbf{N/A for both restrictions}. N/A marks a property of the matched-magnitude comparison rule, not a coverage gap of the published method itself.

\textbf{Finding 2 --- Aggregate directional advantage.} Across the 36 comparable cells, our family leads the matched 1D restriction in \textbf{33 cells (91.7\%)}. Three single-cell losses are observed; only $HG_{4,d+2}$ vs.\ $\todial{}(+2)$ on hate falls outside the reference band, and Section~\ref{sec:granularity} attributes it to a granularity mismatch within our family.\footnote{Cells share training data, the same student/teacher pair, and overlapping configuration substructure (the $H$ and $HG$ families share the $G_2$ or $G_5$ base), so we treat these counts as aggregate directional evidence rather than independent hypothesis tests.} Section~\ref{sec:seeds} shows that the magnitudes of individual cells in this count are not stable across seeds, which is why we report the count and not the per-cell gaps.

\textbf{Finding 3 --- Multi-coefficient combinations are where the largest sweep gains occur.} Single-channel restrictions (the $G$ family with $d=0$) match or marginally exceed the corresponding 1D restrictions but do not reach the strongest cells of our family; only multi-coefficient configurations ($H$, $HG_5$) attain both the family-internal best accuracy and the largest $\Delta$ over the aligned restrictions. Together with the 26-cell isolation of Section~\ref{sec:mean-match}, the aggregate direction is positive in both comparisons.

\section{Seed Robustness and a Third Task}
\label{sec:seeds}

\begin{table*}[t]
\centering
\small
\setlength{\tabcolsep}{5pt}
\begin{tabular}{llccccc}
\toprule
Axis & Task & Comparison & 1-seed $\Delta$ & 3-seed $\Delta$ (mean $\pm$ SD) & 95\% CI & $p$ \\
\midrule
\multirow{6}{*}{\shortstack[l]{Restriction\\(Section~\ref{sec:vs-prior})}}
 & hate      & $H_1$ vs.\ \eopdal{}$(-4)$        & $+4.21$ & $+1.73 \pm 2.18$ & $[-3.69, +7.15]$ & 0.30 \\
 & hate      & $HG_{4,d+2}$ vs.\ \todial{}$(+2)$  & $-1.78$ & $-0.72 \pm 0.92$ & $[-3.00, +1.57]$ & 0.31 \\
 & emotion   & $HG_{5,d+4}$ vs.\ \eopdal{}$(-4)$  & $+1.97$ & $+0.63 \pm 1.27$ & $[-2.52, +3.79]$ & 0.48 \\
 & offensive & $H_1$ vs.\ \eopdal{}$(-4)$        & $+2.33$ & $+0.85 \pm 1.52$ & $[-2.91, +4.62]$ & 0.43 \\
 & offensive & $H_1$ vs.\ \todial{}$(+2)$        & $+1.40$ & $+0.93 \pm 1.13$ & $[-1.86, +3.73]$ & 0.29 \\
 & offensive & $G_2$ vs.\ \eopdal{}$(-4)$        & $+1.40$ & $+0.78 \pm 1.08$ & $[-1.91, +3.46]$ & 0.34 \\
\midrule
\multirow{3}{*}{\shortstack[l]{Isolation\\(Section~\ref{sec:mean-match})}}
 & hate      & $G_4$ vs.\ static $\bar{\lambda}{=}0.556$      & $+3.84$ & $+1.43 \pm 3.09$ & $[-6.24, +9.09]$ & 0.51 \\
 & emotion   & $HG_{5,d+4}$ vs.\ static $\bar{\lambda}{=}0.388$ & $+1.41$ & $+0.47 \pm 0.83$ & $[-1.58, +2.52]$ & 0.43 \\
 & offensive & $H_1$ vs.\ static $\bar{\lambda}{=}0.574$      & $+2.33$ & $+1.28 \pm 1.91$ & $[-3.48, +6.04]$ & 0.37 \\
\bottomrule
\end{tabular}
\caption{Three-seed paired replication of the nine headline comparisons, on both evaluation axes and three tasks. Gaps are in percentage points of accuracy and paired within seed (both sides retrained per seed, seeds $42/2/3$); $p$ from a two-sided paired $t$-test with $n=3$; CIs are $t$-based and necessarily wide at $n=3$, and are reported as uncertainty summaries rather than confirmatory tests. Every three-seed mean is smaller in magnitude than its single-seed counterpart; eight of nine remain directionally positive; none is individually significant. Per-seed accuracies are in Appendix~\ref{app:seeds}.}
\label{tab:3seed}
\end{table*}

The sweeps above use one seed per run, which supports aggregate directional counts but not claims about individual cell gaps. To quantify this, we selected --- after inspecting the single-seed sweep, and fixed before running any additional seed --- the headline comparisons it singles out along \emph{both} axes, and replicated each with three seeds ($42$, $2$, $3$), retraining \emph{both} sides so every gap is paired within seed. We also added a third task, TweetEval offensive (binary, $n_{\mathrm{test}}=860$), trained with the identical protocol (200 steps, same batch size, learning rate, and teacher/student pair), and applied the same three-seed replication there.

\paragraph{Third task.} On offensive, a single-seed sweep of the same family selects $G_2 = (4, 4, -1.5, 0)$ as the strongest $d=0$ configuration and $H_1 = (4, 4, -1.5, +2)$ as the strongest overall --- the same $H$ family that is strongest on hate. We replicate three restriction-axis comparisons ($H_1$ and $G_2$ against \eopdal{}$(-4)$, $H_1$ against \todial{}$(+2)$) and one isolation-axis comparison ($H_1$ against its mean-matched static baseline at $\bar{\lambda} = 0.574$).

\paragraph{Results.}
Table~\ref{tab:3seed} reports all nine replications. Two observations follow.

First, \textbf{every three-seed mean is smaller in magnitude than its single-seed estimate}, by a factor of roughly $1.5$--$3.1$: the single-seed point estimates were optimistic, and we adopt the three-seed means as the more reliable effect sizes. No comparison is significant at $n=3$ ($p \in [0.29, 0.51]$), as expected given across-seed SDs of $0.2$--$1.8$pp on splits of $860$--$2970$ examples; we therefore make no per-cell significance claims anywhere in this paper.

Second, \textbf{eight of the nine replicated gaps remain directionally positive}, on both axes and on all three tasks, including the newly added offensive task. The one negative entry is the hate $HG_{4,d+2}$ cell that Section~\ref{sec:granularity} flagged as a granularity mismatch, and it shrinks from $-1.78$pp to $-0.72 \pm 0.92$pp --- i.e.\ the single documented counterexample in the sweep also moves inside the reference band, so we no longer describe it as a meaningful reversal.

\paragraph{Two distinct sources of uncertainty.}
The reference bands in Sections~\ref{sec:main}--\ref{sec:vs-prior} are \emph{test-set sampling} SEs ($\sigma \approx 1.1$pp on emotion, $0.9$pp on hate, $1.45$pp on offensive), i.e.\ how far a fixed model's measured accuracy can move on a finite split. Across-seed \emph{training} variability is a separate quantity that the three-seed runs let us estimate for the first time (Appendix~\ref{app:seeds}): $0.2$--$1.8$pp. Notably the mean-matched static baselines are far more seed-stable (SD $=0.20$pp on emotion and offensive) than the dynamic configurations ($0.9$--$1.8$pp): per-token gating amplifies sensitivity to initialization and data order. The $\pm 1\sigma$ bands in Figure~\ref{fig:mean-match} are therefore heuristic references for a single model's accuracy, not confidence intervals for the plotted differences.

\paragraph{What this implies for the aggregate counts.}
The nine replications are directionally consistent with the 33/36 and 19/26 counts but do not independently validate them: the counts remain summaries of correlated single-seed sweeps, and multi-seed replication of the full grid was beyond our compute budget. The paper's empirical claim is therefore the group-level one: within short-output classification OPD on a single Qwen3-32B/4B pair, multi-coefficient per-token gating is directionally ahead of both the matched 1D restrictions and the effective-KL-matched static baselines, with headline magnitudes of roughly $0.5$--$1.7$pp and wide uncertainty.

\section{Conclusion}
\label{sec:conclusion}

We introduced a four-coefficient parametric family $\lamt = \sigma(a \hpt + b \ux + c + d \gapt)$ for per-token KL gating in on-policy distillation, and showed that direction-aligned proxies of \eopd{} and \todi{} are single-channel points inside it. The parameterization supports three contributions: (i) a \emph{controlled comparison} of two previously incomparable gating designs at matched magnitude inside one implementation, where the full family leads the matched 1D restriction in 33 of 36 cells, with headline magnitudes of roughly $0.5$--$1.7$pp and wide intervals over three seeds; (ii) \emph{multi-coefficient composition} extending the union of the two restrictions --- hate's strongest configuration uses the \ux{} channel, which neither proxy can express; and (iii) an \emph{isolation protocol} training a static baseline at each configuration's emergent effective KL ratio, under which dynamic gating leads in 19 of 26 cells and the three per-task headline pairs stay positive across seeds. All conclusions are scoped to short-output classification OPD with one Qwen3-32B/4B pair and stated at the group level; the family is a coordinate system for comparing gating designs, not a turnkey method.

\label{pg:concend}
\section*{Limitations}
\label{sec:limit}

\textbf{Per-cell effects are not established; only group-level direction is.}
The 13-configuration grid, its aligned restrictions, and the 26-cell mean-match isolation were each run with a single seed. We replicated the nine headline comparisons that this sweep singles out with three seeds on both sides (Section~\ref{sec:seeds}); all three-seed means came out smaller than the single-seed estimates and none was significant at $n=3$ ($p \ge 0.29$). We therefore make no claim about any individual configuration's advantage, including the largest gaps in the sweep, and the aggregate counts (33/36, 19/26) should be read as exploratory summaries of correlated sweeps rather than as statistical tests. A full multi-seed replication of the grid was outside our compute budget and remains the most important missing piece of evidence.

\textbf{Aligned restrictions are proxies, not reproductions.}
Our comparators are 1D restrictions inside our own convex-mixture implementation. As enumerated in Section~\ref{sec:proxy-caveat}, the \todial{} restriction substitutes a prob-diff gap for \todi{}'s log-ratio gate and drops its stop-gradient, and the \eopdal{} restriction replaces \eopd{}'s additive hard switch with a convex relaxation; neither uses the $\beta$ tuned in the original work. Results support claims about restrictions of our family under matched magnitude, not about the published systems.

\textbf{Matched magnitude rather than best-tuned $\beta$.}
Matching signal magnitude is a deliberate control that isolates gating \emph{structure} from the confound of differing average FKL/RKL ratios, in the same spirit as the mean-matched static baseline. It does not answer the complementary question of how the family compares to each prior design at its own best-tuned $\beta$ over a wide range. Table~\ref{tab:best-observed} gives only a partial best-observed view inside our existing sweep.

\textbf{Task and model scope.}
All three tasks are short-text, short-output classification (responses of 1--3 tokens) from TweetEval, with a single Qwen3-32B/4B teacher--student pair. We chose this regime deliberately: with responses of 1--3 tokens, classical exposure-bias arguments for per-token gating (long-horizon train/test drift) are minimal, so a gain must come from token-level signal selection. But the regime is also narrow. Long-form generation, instruction following, and reasoning --- where per-token dynamics, sequence length, and reward structure differ substantially --- are outside the scope of this paper, as are other architectures and scales; we make no claims about them, and our conclusions should not be extrapolated to those settings.

\textbf{Coefficient selection requires a grid search.}
The family is a linear combination of signals used by prior work, and we select $(a,b,c,d)$ by manual design plus grid search, with no principled automatic criterion and no theory predicting which coefficients a task will prefer. This limits practical use: the parameterization is best viewed as a shared coordinate system for analysis and comparison rather than as a turnkey method.

\textbf{Reproducibility without a code release.}
Our gating implementation is embedded in internal training infrastructure that we are unable to release, so our results cannot be reproduced by running our code, and an independent re-implementation is required. To make that feasible we fully specify: the gating function and its three signals (Equation~\ref{eq:family}, Section~\ref{sec:coef}); the exact $(a,b,c,d)$ values of every configuration we train, including the aligned restrictions and the mean-matched static baselines (Section~\ref{sec:main}, Sections~\ref{sec:vs-prior}--\ref{sec:seeds}); all training hyperparameters, seeds, decoding settings, and the output parser (Appendix~\ref{app:repro}); and per-seed accuracies for every run behind Table~\ref{tab:3seed} (Appendix~\ref{app:seeds}). Absolute accuracies are nevertheless sensitive to the prompt template and parser, and --- as Section~\ref{sec:seeds} shows --- to the training seed, so a re-implementation should be expected to reproduce the direction and rough magnitude of the reported effects rather than exact numbers. This is a genuine limitation on the verifiability of our results.

\label{pg:limend}   

\section*{Acknowledgments}

We thank the anonymous reviewers and the area chair for detailed and constructive feedback; in particular, their insistence on multi-seed evidence directly produced the replication study in Section~\ref{sec:seeds} and led us to retire several per-cell claims from the submitted version. We also thank our colleagues for infrastructure and compute support.

\bibliography{references_verified}

\appendix

\section{On-Policy Distillation Background and Infrastructure}
\label{app:repro}

\paragraph{Background.}
Standard distillation is \emph{offline}: the teacher generates pre-computed responses, and the student trains on the static (prompt, response) pairs. On-policy distillation (OPD) \citep{agarwal2024onpolicy} instead has the student generate responses during training, with the teacher computing per-token target probabilities on-the-fly. This addresses the train--test distribution shift inherent in offline KD: the student learns to refine its own generation distribution rather than mimic a fixed teacher target.

\paragraph{Software stack.}
Our implementation builds on the verl framework \citep{sheng2024hybridflow}. The student (Qwen3-4B) generates responses via vLLM \citep{kwon2023efficient}, TP=2. A separate teacher service (Qwen3-32B \citep{qwen3}, TP=2) is a vLLM instance that computes per-token teacher probabilities on the student's rollout tokens, with ZMQ-based message passing between teacher and student processes. Student backpropagation uses Megatron-LM \citep{shoeybi2019megatronlm} with TP=2 and DP=3. This stack is internal and is not released; the specification below is intended to be sufficient for re-implementation on top of any on-policy distillation trainer that exposes per-token teacher log-probabilities.

\paragraph{Training hyperparameters.}
Student Qwen3-4B, teacher Qwen3-32B; learning rate $10^{-6}$, batch size 72; 100 training steps on emotion, 200 on hate and offensive; rkl\_ratio $=0.5$ for the static $\lambda{=}0.5$ baseline and $\mathbb{E}[\lamt]$ for the mean-matched static baselines; adv\_estimator $=$ reinforce\_plus\_plus; reward $=$ constant 0 (distillation-only); 6$\times$L20Y 80GB.

\paragraph{Seeds.}
The single-seed sweeps use seed 42. The three-seed replications in Section~\ref{sec:seeds} use seeds $\{42, 2, 3\}$; the seed is injected into Megatron weight initialization, the data loader, and data shuffling, and both sides of every reported gap are retrained under the same seed so that all gaps are paired. One hate configuration ($HG_{5,d+4}$) uses seeds $\{2, 3, 42\}$ from a separate replication batch.

\paragraph{Evaluation.}
TweetEval held-out evaluation split, scored as classification accuracy on the parsed label. Prompts are single-turn and instruct the model to emit one line of strict JSON, \texttt{\{"label": "<option$_1$|...|option$_k$>"\}}, with the label set enumerated and briefly defined in the prompt and two format examples appended; no in-context task examples are given. Decoding is greedy (temperature 0) with a 32-token cap and Qwen3 thinking disabled, so responses are 1--3 tokens of label text. Parsing takes the first \texttt{"label": "..."} match; if absent, it scans the first 200 characters for a legal label string, preferring the longest match so that e.g.\ \texttt{non\_ironic} is not truncated to \texttt{ironic}; unparseable responses count as errors. For the restriction comparison, 7 unique aligned configurations per task (4 \todial{} with $d \in \{\pm 2, \pm 4\}$ and 3 \eopdal{} with $|a| \in \{2, 4, 8\}$) were trained; checkpoints were evaluated with vLLM at TP=2.

\paragraph{Data.}
TweetEval \citep{barbieri2020tweeteval} emotion (4-way, $n_{\mathrm{train}}=3257$, $n_{\mathrm{test}}=1421$); hate (binary, $n_{\mathrm{train}}=8993$, $n_{\mathrm{test}}=2970$); offensive (binary, $n_{\mathrm{train}}=11916$, $n_{\mathrm{test}}=860$).

\section{Three-Seed Replication Details}
\label{app:seeds}

Table~\ref{tab:3seed-percfg} lists the per-seed accuracies and across-seed SDs behind Table~\ref{tab:3seed}. Two patterns are worth recording. First, across-seed SD varies by an order of magnitude between configurations ($0.20$--$1.82$pp), and is largest exactly for the configurations that produced the largest single-seed gaps (hate $G_4$, hate $H_1$, offensive $H_1$) --- a selection effect that explains why single-seed headline numbers were optimistic. Second, the mean-matched static baselines are the most seed-stable runs in the table (SD $=0.20$pp on emotion and offensive), so the width of the dynamic-vs-static gaps in Table~\ref{tab:3seed} is driven almost entirely by variability on the dynamic side.

\begin{table}[t]
\centering
\small
\setlength{\tabcolsep}{3.5pt}
\begin{tabular}{llccc}
\toprule
Task & Config & s42 & s2 & s3 \\
\midrule
\multirow{6}{*}{\shortstack[l]{emotion\\($n{=}1421$)}}
 & $G_4$                & 0.7734 & 0.7685 & 0.7671 \\
 & $G_5$                & 0.7748 & 0.7635 & 0.7847 \\
 & $H_2$                & 0.7847 & 0.7734 & 0.7706 \\
 & $HG_{5,d+4}$          & 0.7868 & 0.7713 & 0.7706 \\
 & \eopdal{}$(-4)$      & 0.7671 & 0.7769 & 0.7657 \\
 & static $0.388$       & 0.7727 & 0.7727 & 0.7692 \\
\midrule
\multirow{8}{*}{\shortstack[l]{hate\\($n{=}2970$)}}
 & $G_4$                & 0.5586 & 0.5239 & 0.5444 \\
 & $G_5$                & 0.5286 & 0.5296 & 0.5529 \\
 & $H_1$                & 0.5549 & 0.5360 & 0.5283 \\
 & $H_2$                & 0.5411 & 0.5488 & 0.5525 \\
 & $HG_{4,d+2}$          & 0.5209 & 0.5357 & 0.5266 \\
 & \eopdal{}$(-4)$      & 0.5128 & 0.5350 & 0.5195 \\
 & \todial{}$(+2)$      & 0.5387 & 0.5377 & 0.5283 \\
 & static $0.556$       & 0.5202 & 0.5444 & 0.5195 \\
\midrule
\multirow{5}{*}{\shortstack[l]{offensive\\($n{=}860$)}}
 & $G_2$                & 0.7721 & 0.7651 & 0.7698 \\
 & $H_1$                & 0.7814 & 0.7791 & 0.7488 \\
 & \eopdal{}$(-4)$      & 0.7581 & 0.7698 & 0.7558 \\
 & \todial{}$(+2)$      & 0.7674 & 0.7616 & 0.7523 \\
 & static $0.574$       & 0.7581 & 0.7547 & 0.7581 \\
\bottomrule
\end{tabular}
\caption{Per-seed held-out accuracy for every run entering Table~\ref{tab:3seed}. Across-seed SDs (pp): emotion $G_4$ 0.33, $G_5$ 1.06, $H_2$ 0.75, $HG_{5,d+4}$ 0.92, \eopdal{} 0.61, static 0.20; hate $G_4$ 1.74, $G_5$ 1.38, $H_1$ 1.37, $H_2$ 0.58, $HG_{4,d+2}$ 0.75, \eopdal{} 1.14, \todial{} 0.57, static 1.42; offensive $G_2$ 0.36, $H_1$ 1.82, \eopdal{} 0.75, \todial{} 0.76, static 0.20.}
\label{tab:3seed-percfg}
\end{table}

\paragraph{Statistics.}
For each comparison we compute the per-seed paired difference and report its mean, SD, a two-sided paired $t$-test, and the $t$-based 95\% CI ($n=3$, $2$ degrees of freedom). With three paired seeds the CI half-width is roughly $2.5$ SDs of the paired differences, so intervals are wide by construction; we report them to communicate uncertainty, not as confirmatory tests. We did not correct for multiple comparisons, since no comparison is significant without correction.

\section{Additional Sweep Views}
\label{app:extra}

\begin{figure}[t]
\centering
\includegraphics[width=\columnwidth]{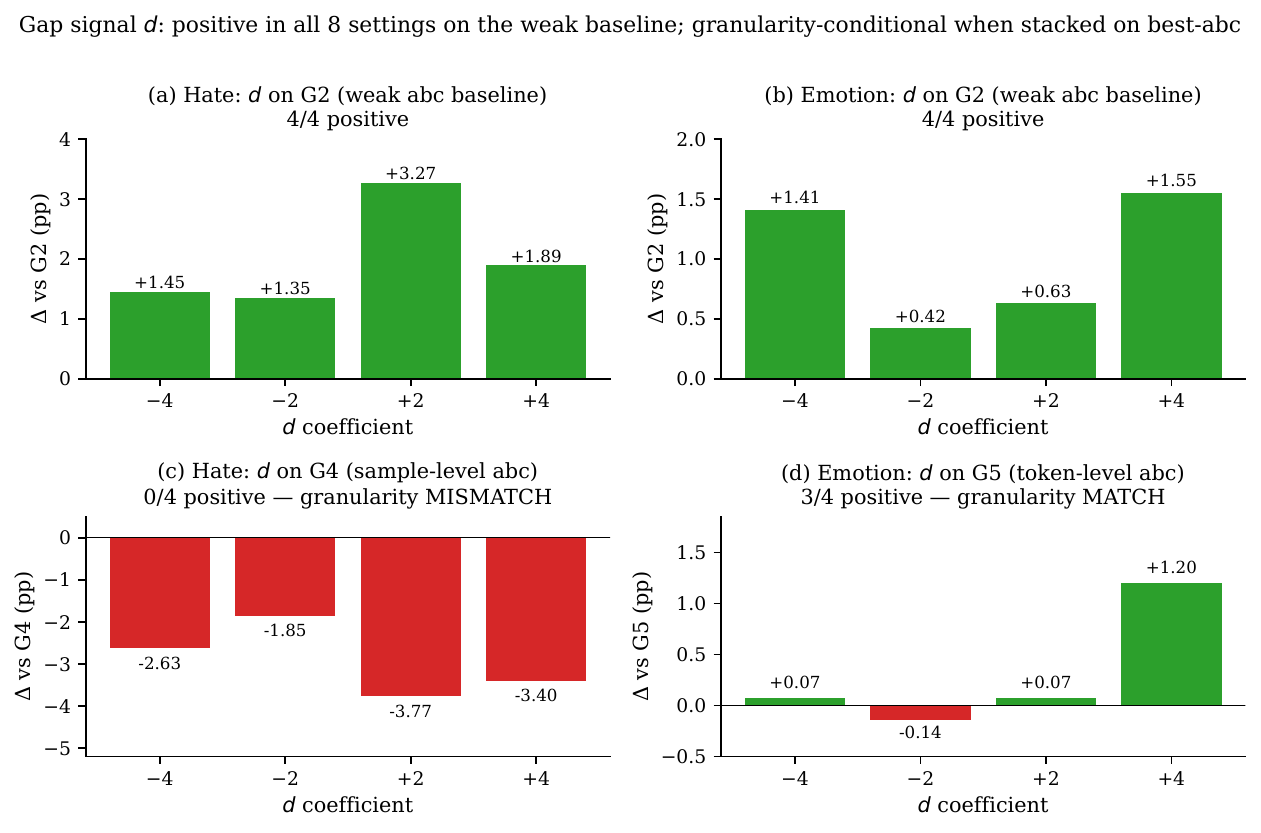}
\caption{Gap signal $d$ (single-seed sweep): positive in all 8 settings on the weak $G_2$ baseline (top row), while stacking on the task-best $abc$ is granularity-conditional --- emotion (token-level, MATCH): 3/4 positive; hate (sample-level, MISMATCH): 0/4 positive. Referenced from Section~\ref{sec:granularity}.}
\label{fig:granularity}
\end{figure}

This appendix holds two views of the single-seed sweep that are referenced from the main text but not needed to follow it. Figure~\ref{fig:granularity} decomposes the effect of the gap coefficient $d$ --- adding it to the weak $G_2$ baseline versus stacking it on each task's best $abc$ configuration (Section~\ref{sec:granularity}). Table~\ref{tab:best-observed} takes the best-observed accuracy of each side within the explored sweep, as a partial alternative to the matched-magnitude comparison of Section~\ref{sec:vs-prior}.

\begin{table}[t]
\centering
\small
\setlength{\tabcolsep}{4pt}
\begin{tabular}{lcccc}
\toprule
Task & Static & \todial{} & \eopdal{} & Ours \\
     & $\lambda{=}0.5$ & best $d$ & best $|a|$ & best \\
\midrule
Emotion & 0.7671 & 0.7734 & 0.7790 & \textbf{0.7868} \\
Hate    & 0.5391 & 0.5387 & 0.5347 & \textbf{0.5586} \\
\bottomrule
\end{tabular}
\caption{Best-observed accuracy per side within our shared single-seed sweep (Section~\ref{sec:best-vs-best}). The \todial{}/\eopdal{} columns take the best of their 1D restrictions ($d \in \{\pm 2, \pm 4\}$, $|a| \in \{2, 4, 8\}$); Ours is the best of the 13-config grid. This is not a per-method hyperparameter optimization, and per-cell magnitudes are seed-sensitive (Section~\ref{sec:seeds}).}
\label{tab:best-observed}
\end{table}

\section{Design Choices Not Available to the Aligned Restrictions}
\label{app:design-choices}

Two structural degrees of freedom exposed by our parameterization are absent from the \todial{} and \eopdal{} 1D restrictions. We record them here for completeness.

\paragraph{Bias coefficient $c$.}
Our family includes a bias term $c$ in the sigmoid input. Both original methods fix $c = 0$ by construction. The best configuration on each task in our sweep uses $c = -1$ ($G_4$ on hate, $G_5$ on emotion), matching Pattern 1 in Section~\ref{sec:coef}; on emotion this contrast is inside the sampling-SE reference band (Section~\ref{sec:taskmirror}). Whether a nonzero $c$ contributes is a question the parameterization makes askable, and our answer for these tasks is descriptive rather than conclusive.

\paragraph{Sign of coefficient $a$.}
\eopd{}'s original semantics fixes $a < 0$ via the ``high entropy $\Rightarrow$ more FKL'' rule, so the \eopdal{} restriction uses $a < 0$ by construction. Emotion's best configuration $G_5$, however, uses $a > 0$ (high entropy $\Rightarrow$ more RKL) --- a regime \eopd{}'s original formulation cannot represent. The sign discrepancy between the \eopdal{} restriction and our family on emotion is therefore an instance of the family extending the union of the two restrictions, not a confound in the matched-magnitude comparison.

\end{document}